\pdfoutput=1

\documentclass[11pt]{article}
\usepackage[preprint]{acl}

\usepackage{times}
\usepackage{latexsym}
\usepackage[T1]{fontenc}
\usepackage[utf8]{inputenc}
\usepackage{microtype}

\usepackage{booktabs}
\usepackage{multirow}
\usepackage{inconsolata}
\usepackage{graphicx}
\usepackage{amsmath}

\title{Logical Reasoning with Outcome Reward Models for Test-Time Scaling}

\author{
 \textbf{Ramya Keerthy Thatikonda\textsuperscript{$\forall$}} \ \ \
 \textbf{Wray Buntine\textsuperscript{$\forall$,$\exists$}} \ \ \
 \textbf{Ehsan Shareghi\textsuperscript{$\forall$}}
\\
\\
 \textsuperscript{$\forall$}Department of Data Science \& AI, Monash University
 \\
 \textsuperscript{$\exists$}College of Engineering and Computer Science, VinUniversity
\\
}

\begin{document}
\maketitle
\begin{abstract}
Logical reasoning is a critical benchmark for evaluating the capabilities of large language models (LLMs), as it reflects their ability to derive valid conclusions from given premises. 
While the combination of test-time scaling with dedicated outcome or process reward models has opened up new avenues to enhance LLMs performance in complex reasoning tasks, this space is under-explored in deductive logical reasoning. We present a set of Outcome Reward Models (ORMs) for deductive reasoning. To train the ORMs we mainly generate data using Chain-of-Thought (CoT) with single and multiple samples. Additionally, we propose a novel tactic to further expand the type of errors covered in the training dataset of the ORM. In particular, we propose an echo generation technique that leverages LLMs’ tendency to reflect incorrect assumptions made in prompts to extract additional training data, covering previously unexplored error types. While a standard CoT chain may contain errors likely to be made by the reasoner, the echo strategy deliberately steers the model toward incorrect reasoning. We show that ORMs trained on CoT and echo-augmented data demonstrate improved performance on the FOLIO, JustLogic, and ProverQA datasets across four different LLMs.\footnote{Code is available at \url{https://github.com/RamyaKeerthy/LogicORM}}
\end{abstract}

\section{Introduction}
\label{sec:intro}

Logical reasoning in large language models~(LLMs) has primarily been studied as a symbolic task using in-context learning \cite{lam2024closerlooklogicalreasoning, logiclm, satlm, linc}, and fine-tuning \cite{strategies-fol, proverqa}.  
Current state-of-the-art techniques in reasoning which bundle  test-time scaling \cite{inference-scaling, scaling-llm} with Process or Outcome Reward Models \cite{math-shepherd, verify-step-by-step, orm-math}, proven effective in math and coding, while remaining heavily underexplored for logical reasoning. This presents a significant opportunity to assess and enhance LLMs' reasoning capabilities using reward models at test-time with text-based reasoning. In this paper we explore logical reasoning with test-time scaling, demonstrating performance gains on three datasets; FOLIO \cite{folio}, ProverQA \cite{proverqa}, and JustLogic \cite{chen2025justlogic} when combined with verification via Outcome Reward Models (ORMs).

Outcome Reward Models (ORMs) enable verification of the entire reasoning sequence by assigning a confidence score to the final output. A central challenge in training ORMs is acquiring high-quality, diverse training data. To address this, we generate multiple Chain-of-Thought~\cite{wei2022chain} reasoning candidates via sampling. Compared to an ORM trained on a single sample per reasoning question, this facilitates more effective re-ranking of solutions during inference.

To further enrich the training data, we leverage the echoing behavior of LLMs - where models tend to align their reasoning with user-provided answers. This phenomenon can introduce hallucinated reasoning, which, when subsequently flagged as incorrect through a second-level filtering process, results in a diverse set of flawed reasoning paths. We show that incorporating these echoed errors into the training set helps the ORM learn to better distinguish between valid and invalid reasoning trajectories during use with test-time scaling.
\begin{figure*}[t]
  \centering
  \includegraphics[scale=0.8, trim=80 225 200 130, clip]{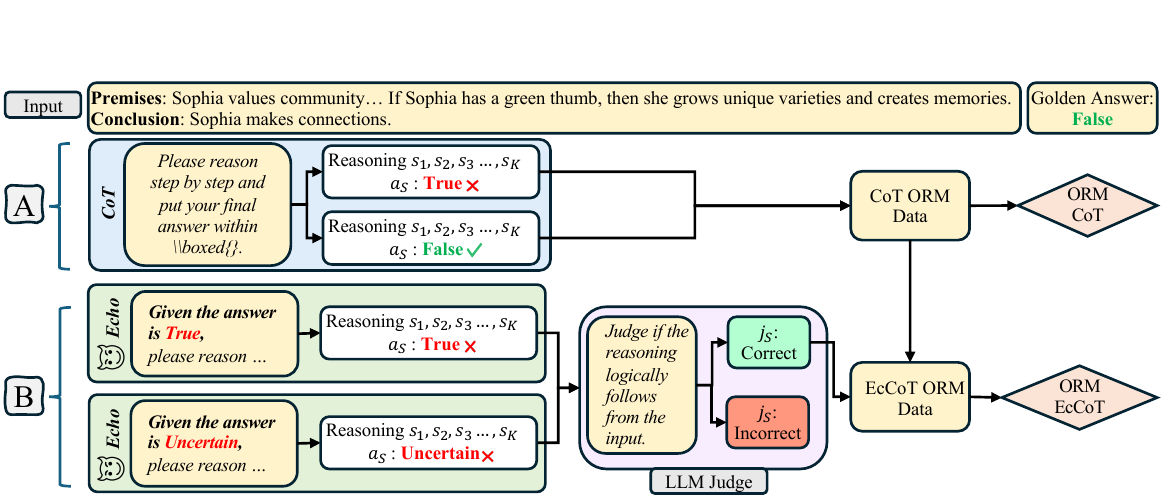}
  \caption{
  Methodology for generating training data for CoT and EcCoT ORMs. \textbf{A.} The LLM is prompted to generate CoT reasoning, and both correct and incorrect answers are used to construct the ORM dataset. \textbf{B.} The LLM is provided with a misleading answer to elicit reasoning. The resulting incorrect reasoning trajectories are then passed back to the LLM for evaluation. If the LLM fails to recognize its error, the trajectory is added to the dataset.}
  \label{fig:method}
\end{figure*}

Our contributions are as follows:
(1) We demonstrate that training ORMs with multiple sampled CoT candidates (per example) significantly improves the reliability of the ORM on deductive reasoning tasks.
(2) We show that augmenting CoT training data with data generated by echoed-error further enhances the resulting ORMs' accuracy.
(3) We compare ORM models on training data size, format, and reward distribution.

\section{Outcome Reward Model for Logic}
\label{sec:methodology}
In outcome supervision, the reward model evaluates the entire reasoning sequence and assigns a final score reflecting the quality of the outcome. During inference, this score can be used to re-rank the candidate solutions generated by a large language model (LLM). Following \citet{verify-step-by-step} recipe for mathematical reasoning, to train an Outcome Reward Model (ORM), each reasoning trace is compared against a gold-standard label: a positive reward is given when the final output matches the correct answer, and a negative reward is assigned otherwise. The ORM learns to map reasoning sequences to these rewards, producing logits that can be used to rank candidate outputs during inference.

We explore two types of data generation strategies for training ORMs:
Outcome supervision on standard Chain-of-Thought (CoT) reasoning, and
Outcome supervision on Echo Chain-of-Thought (EcCoT) - a variant that incorporates the LLM’s echoing behavior to produce additional reasoning paths. See Figure~\ref{fig:method} for an overview.

\paragraph{CoT ORM Data Generation.}
For standard CoT data generation, we prompt an LLM with the context and question to generate step-by-step reasoning, with the final answer appended at the end. The LLM generates multiple reasoning candidates for a given question. These candidates are then parsed into individual CoT traces. We used "\texttt{Please reason step by step, and put your final answer within \textbackslash\textbackslash{boxed}\{\}}" as prompt.

Each candidate is labeled with a reward: positive if the final answer matches the gold label, and negative otherwise. This data is used to train a second LLM as a classifier that predicts the reward score based on the reasoning sequence. Our approach builds on the ideas presented in \citet{math-shepherd}, but differs by applying outcome supervision rather than process supervision, and by using a straightforward automated annotation to label the data. 

\paragraph{EcCoT ORM Data Generation.} 
LLMs often exhibit a tendency to follow user-provided answers uncritically, sometimes hallucinating or forcefully aligning their reasoning to fit an incorrect answer. We exploit this behavior to generate challenging negative examples.

In contrast to \citet{star-paper}, we prompt the LLM with the instruction "\texttt{Given the answer is True, please reason step by step, and put your final answer within \textbackslash\textbackslash{boxed}\{\}}" for all reasoning questions. This coerces the model into producing reasoning that unjustifiably supports the provided answer (where we deliberately provided an incorrect answer, e.g. "True" where the correct answer was "False") resulting in flawed reasoning trajectories. These incorrect yet plausible reasoning sequences, referred to as echoes, are valuable for training ORMs to penalize invalid rationales.

While echoing can encourage LLMs to commit to flawed reasoning paths, we further filtered the collected echoes by prompting the LLM (i.e., "\texttt{Judge if the reasoning logically follows from the input; respond only with Correct or Incorrect.}") itself to evaluate whether the reasoning trajectories were correct. We discarded the echoed examples that the LLM identified as incorrect, as these were deemed too obvious and unlikely to occur during inference. The remaining echoes~(those involving incorrect reasoning and not easily recognized as such) were retained and added to the training data alongside the CoT examples.

The statistics for CoT and EcCoT, together with sample prompt outputs, are provided in Appendix~\ref{sec:sample}.

\section{Experimental Setup}
\label{sec:experiments}

The ORMs utilized in this study were trained using the Qwen 2.5 7B Instruct models \cite{hui2024qwen2} on a single A100 GPU for three epochs, with a batch size of 64 and a learning rate of 5 × $10^{-4}$. Training was conducted using a LoRA-based PEFT configuration \cite{hu2022lora}. For result annotation, we apply a step tag <extra\_0>, with positive and negative outcomes indicated by `+' and `-', respectively following \citet{zhang2025lessons}.

\paragraph{Logical Reasoning Datasets.}
We use ProverQA~\cite{proverqa}, JustLogic~\cite{chen2025justlogic} and FOLIO~\cite{folio} training sets to generate data for training ORMs, sampling 8, 8 and 10 reasoning candidates per instance, respectively. The ProverQA training set comprises of 5,000 logical reasoning questions across three difficulty levels (easy, medium, and hard) with additional noise-premise variations. JustLogic training set consists of 4900 synthetically generated logical reasoning questions with difficulty spanning across 7 reasoning depths. FOLIO contains around 1,000 human-annotated deductive reasoning questions. All datasets involve deriving a conclusion based on a given premise. A well-known example of deductive reasoning is: 
\begin{quote}
Premises: All humans are mortal. Jack is a human.
Conclusion: Jack is mortal. Labels: True/False/Uncertain
\end{quote}

\subsection{Data Generator}
We refer to the the model used to generate the  training data as the \emph{generator}. We employ two types of generators with varying sample size.

\paragraph{Qwen2.5 Data Generator (\emph{Qwen-as-a-Generator}).}
Our initial experiments aim to validate the hypothesis that training with 8 samples from a generator improves ORM performance compared to training with a single sample. We use the Qwen2.5-7B Instruct model to generate both small (1-generator sample) and large (8-generator samples) CoT datasets on ProverQA, resulting in two models: ORM-CoT$^{small}$ and ORM-CoT$^{large}$. We further augment the large CoT dataset with Echo Chain-of-Thought (EcCoT) data, using the same LLM, to produce ORM-EcCoT$^{large}$. 

\paragraph{GPT-4o Data Generator (\emph{GPT4o-as-a-Generator}).}
We run a set of experiments by using GPT-4o to generate CoT and EcCoT data for ProverQA, FOLIO and JustLogic\footnote{We apply resampling to address the large volume of echoes generated by GPT-4o for the JustLogic dataset. See Appendix~\ref{app:sampling} for details of the sampling and analysis.}. This results in six ORM variants, two for each dataset.

For details on the statistics of the data generated for ORMs, see Appendix~\ref{sec:sample}.
\subsection{Reasoner}
We refer to the model that generates reasoning samples at inference time using test data as the \emph{reasoner}.
The reasoner generates $N$ samples per query, which are evaluated using Best-of-N sampling \cite{verify-step-by-step} where out of the N samples, the sample receiving the highest score from the ORM is selected. We first evaluate the ORM models from \emph{Qwen-as-a-Generator} using two reasoners, Qwen2.5-7B Instruct, and GPT-4o, on the ProverQA (hard) test set.  Encouraged by these results, we extend our evaluation to the FOLIO and JustLogic datasets, incorporating additional reasoners (LLaMA-3.1-8B \cite{grattafiori2024llama} and Qwen3-8B \cite{yang2025qwen3technicalreport}) and leveraging ORM variants trained with data from \emph{GPT4o-as-a-Generator}.

\section{Results and Discussion}
\label{sec:results}
\begin{figure*}[t]
  \centering
  \includegraphics[width=\linewidth, trim=15 70 15 70, clip]{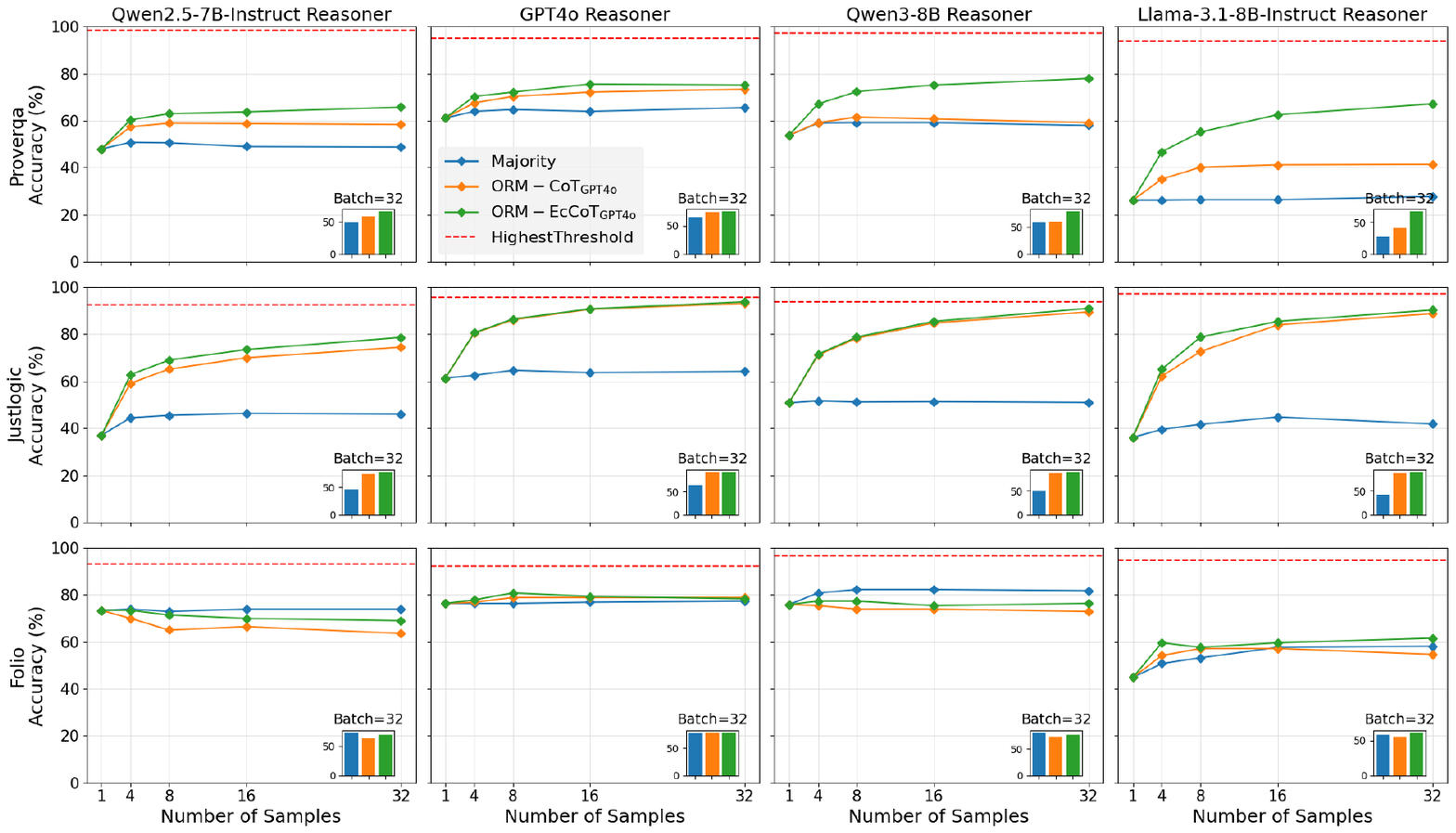}
  \caption{Performance of the ORM trained on GPT-4o-generated data using Chain-of-Thought (CoT) and Echoed formats for ProverQA, JustLogic, and FOLIO. ORM inference was performed using the Best-of-N method for both CoT and Echoed generations. The red dashed line denotes the maximum achievable accuracy assuming at least one correct rationale among the $N$ sampled responses. }
  \label{fig:cotvsechomain}
\end{figure*}

\subsection{Preliminary Experiments}
We first establish a baseline to justify the selection of ORM training data parameters, such as generator sample size (small vs large), data generation method (CoT vs EcCoT), and generator model choice (GPT4o vs Qwen).
\begin{figure}[t]
  \centering
  \includegraphics[width=\linewidth, trim=15 135 15 132, clip]{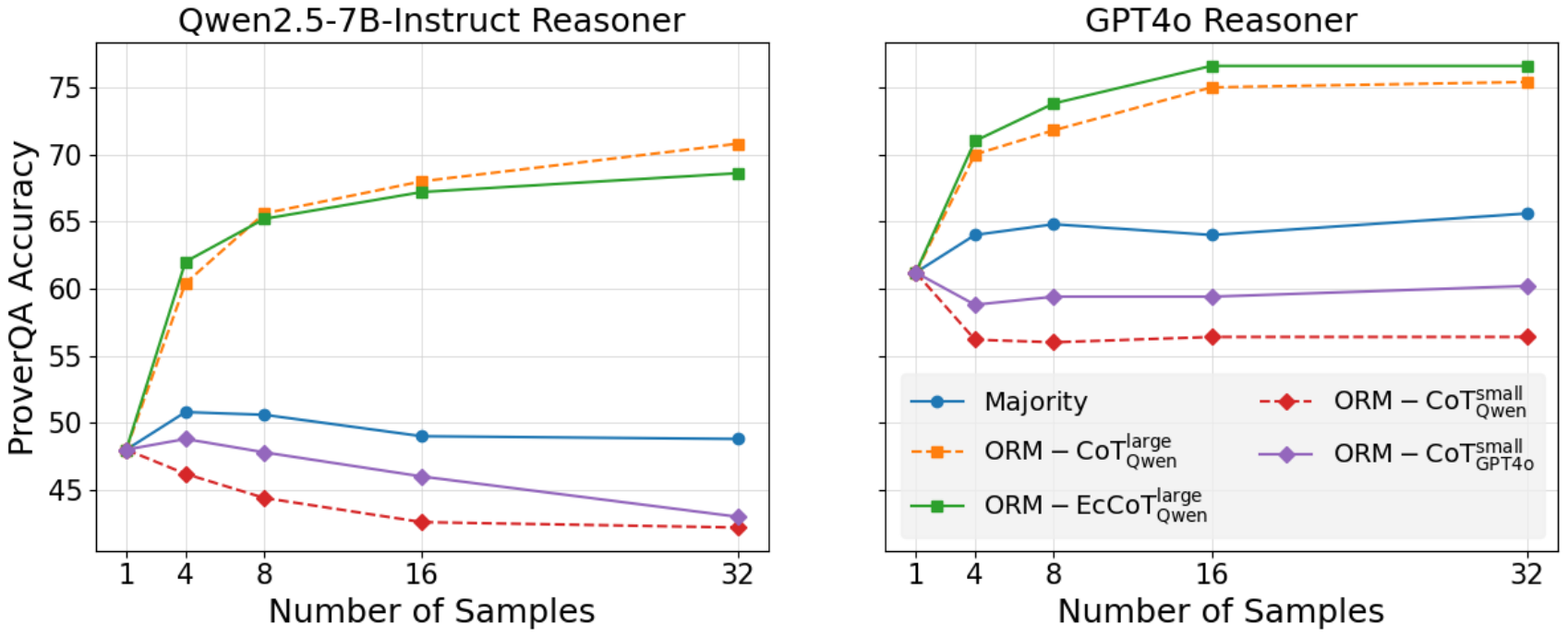}
  \caption{Performance comparison of ORM-trained models based on training data size, data generator, and reasoning generator for ProverQA dataset.}
  \label{fig:cases}
  \vspace{-4mm}
\end{figure}
\paragraph{Sample Size.}
Figure~\ref{fig:cases} illustrates the performance variation of an ORM trained using CoT data generated by the Qwen model. The primary difference across configurations lies in the number of samples generated per question. As the number of samples increases from 1 to 8, the ORM's performance surpasses that of majority voting. This result suggests that generating multiple samples per question, while keeping the training set and questions fixed, can yield more effective training data compared to relying on a single sample.

\paragraph{Augmenting Echo CoT.}
Figure~\ref{fig:cases} presents the performance comparison between EcCoT and standard CoT using 8 generated samples. Incorporating Echo data enhances the ORM and consequently improves the performance of the GPT-4o reasoner, while the improvement is less pronounced for the Qwen reasoner. These results highlight the potential value of Echo, motivating further investigation into its effects.

\paragraph{Generator Model.}
Figure~\ref{fig:cases} examines the differences between generator models. A single-sample CoT generated by Qwen and GPT-4o reveals noticeable variation, with GPT-4o consistently outperforming Qwen. These results motivate the selection of GPT-4o as the preferred generator model (explored next).

\subsection{Main Results and Analysis}
Based on the observations from above, we opt for GPT4o as a generator and train ORMs for ProverQA, JustLogic, and FOLIO datasets.

\paragraph{ORM with Deductive Reasoning.}
We use two ORMs: CoT and EcCoT for all the benchmarks. These models are evaluated using test-time scaling outputs generated by four different reasoners. 

For FOLIO, the relatively small training set~($1{,}000$ records) is reflected in the performance trends shown in Figure~\ref{fig:cotvsechomain}. Majority voting yields a wide accuracy range, from 40\% to 80\%, depending on the model. Qwen, and GPT family of models demonstrate strong performance on FOLIO without requiring additional verification. In contrast, LLaMA begins with lower majority voting accuracy, underscoring the potential benefits of ORM methods on this dataset. While ORM-CoT improves upon majority voting with up to 16 samples, its performance declines at higher sample sizes. In comparison, ORM-EcCoT consistently outperforms both majority voting and ORM-CoT across all sample sizes, demonstrating its robustness and effectiveness on smaller datasets like FOLIO.

The large volume of training data in the ProverQA and JustLogic datasets contributed to notable performance gains for both the CoT and Echo CoT models. These models consistently outperformed the majority vote baseline, with the most substantial improvements observed on logical reasoning benchmarks. Notably, ORMs trained with EcCoT consistently outperformed those trained with CoT on ProverQA, reinforcing our hypothesis that incorrect rationales echoed by the LLM can be leveraged effectively. 

\paragraph{Highest Threshold (HT).} To further analyze performance limits, we measure HT, representing the maximum achievable accuracy assuming at least one correct rationale exists among the $N$ sampled responses. For JustLogic, the HT is nearly equivalent to the performance of the CoT-trained ORM, which explains the limited impact of EcCoT in this case, as the CoT ORM has already reached a performance ceiling. In contrast, the other two benchmarks demonstrate a noticeable gap between ORM performance and their respective HT values. This indicates untapped potential and suggests that a well-designed verification mechanism alone could drive substantial gains in reasoning accuracy, making it a promising direction for future research.

\begin{figure}[t]
  \centering
  \includegraphics[width=\linewidth, trim=150 165 150 165, clip]{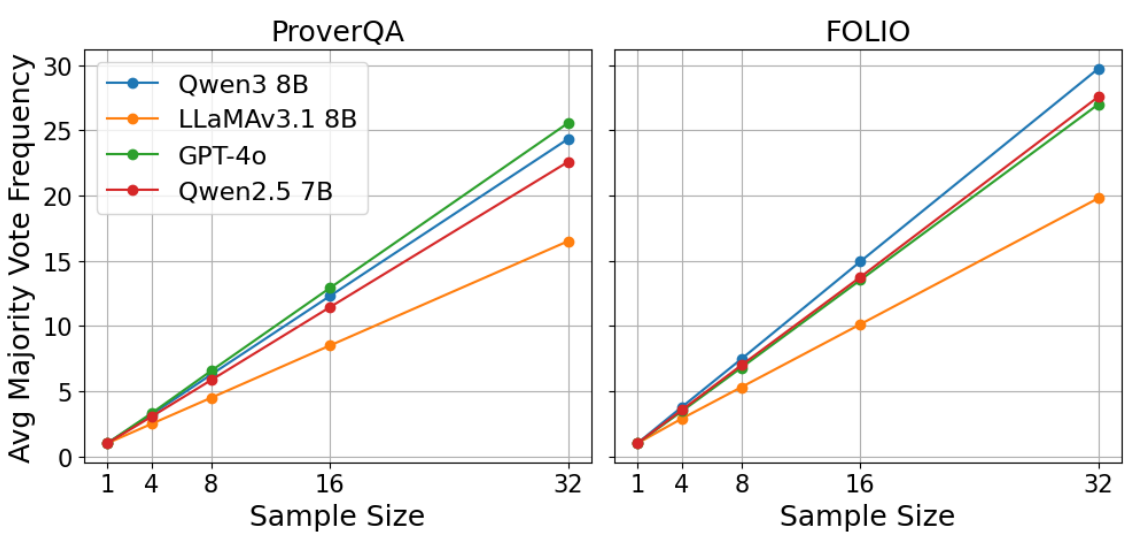}
  \caption{Average majority vote performance across varying sample sizes and benchmarks for different LLM reasoners. Notably, FOLIO achieves close to 32 correct answers on average, suggesting limited room for ORM-based improvement due to the high confidence and consistency of the reasoners.}
  \label{fig:majvot}
  \vspace{-4mm}
\end{figure}

\paragraph{Majority Vote Frequency.}
To address discrepancies observed in the results, we analyzed the majority vote frequency across different sample sizes. This metric captures the average number of correct rationales generated by the reasoner across 
$N$ samples. Figure~\ref{fig:majvot} presents the majority vote frequency for the three benchmarks, which we directly relate to the performance trends shown in Figure~\ref{fig:cotvsechomain}. In the case of FOLIO, the reasoning paths are correct in nearly 90\% of the samples, suggesting that ORM has limited room for improvement over the majority vote. In contrast, the other two benchmarks show a lower proportion of correct answers per sample, providing ORM with a more diverse set of reasoning paths to select from.

In addition to these results, we present two ablation studies (Appendix~\ref{sec:ablation}). The first examines the effect of using EcCoT versus CoT with larger sample sizes. The second analyzes the impact of ORM on reasoners of different sizes (i.e., Gemma3- 1, 4, and 12B variants \cite{team2025gemma}), highlighting the benefits of reward models for smaller language models in test-time settings.

\section{Conclusion}
\label{sec:conclusion}
In this work, we propose the use of outcome reward models~(ORMs) supervised on the final outputs of reasoning paths as a framework for exploring test-time scaling in text-based reasoning. We present a diverse set of ORMs trained on varying model sizes and configurations, and evaluate their performance on three logical reasoning benchmarks - FOLIO, JustLogic, and ProverQA. To enrich training data, we advocate for sampling multiple Chain-of-Thought (CoT) responses and incorporating Echo-based augmentations. Our results provide strong empirical support for this approach. Future work may explore process reward models to assess the correctness of intermediate reasoning steps.

\section{Limitations}
Considering the use of GPT-4o models, we acknowledge the inherent uncertainty associated with data generation. While these models are capable of producing high-quality outputs, they remain susceptible to hallucinations, inconsistencies, and spurious correlations, especially when prompted to generate complex reasoning chains. 

In this work, we explored Outcome supervision, focusing only on the correctness of the final answer without explicitly verifying the validity or faithfulness of the entire reasoning process. This approach can overlook intermediate errors that may still lead to the correct final answer, thus introducing a risk of reinforcing flawed or superficial reasoning patterns during training. 
\bibliography{custom}

\appendix
\label{sec:appendix}
\section{Data Generation}
\label{sec:sample}
For CoT (Chain-of-Thought) data generation, we use a single prompt: "\texttt{Please reason step by step, and put your final answer within \textbackslash\textbackslash{boxed}\{\}}". This elicits a reasoning path with the final answer appearing at the end. For Echo generation, we slightly modify the prompt to: "\texttt{Given the answer is True, please reason step by step, and put your final answer within \textbackslash\textbackslash{boxed}\{\}}". In Tables~\ref{tab:qwen-proverqa}-\ref{tab:justlogic_datagen}, the Echo column displays generations for each label, regardless of the ground truth. When prompted to Echo, the number of incorrect outputs increases, suggesting not only echoed reasoning but also the presence of other ambiguous or flawed reasoning paths.

A sample for comparison of reasoning outcomes for a echoed prompt and regular CoT is presented in Table~\ref{tab:echo-vs-cot}.

\begin{table}[htbp]
\centering
\small
\setlength{\tabcolsep}{4pt} 
\begin{tabular}{llccc}
\toprule
\textbf{Mode} & \textbf{Echo} & \textbf{Total} & \textbf{Correct} & \textbf{Incorrect} \\
\midrule
CoT 0-shot & -- & 39919 & 29962 \raisebox{0.35ex}{\tiny(75\%)} & 9957 \raisebox{0.35ex}{\tiny(25\%)} \\
\midrule
\multirow{3}{*}{Echo 0-shot} 
  & True      & 39935 & 29887 \raisebox{0.35ex}{\tiny(75\%)} & 10048 \raisebox{0.35ex}{\tiny(25\%)} \\
  & False     & 39925 & 29081 \raisebox{0.35ex}{\tiny(73\%)} & 10844 \raisebox{0.35ex}{\tiny(27\%)} \\
  & Uncertain & 39961 & 27291 \raisebox{0.35ex}{\tiny(68\%)} & 12670 \raisebox{0.35ex}{\tiny(32\%)} \\
\midrule
    Echo-CoT & -- & 46469 & 29962 \raisebox{0.35ex}{\tiny(64\%)}& 16507 \raisebox{0.35ex}{\tiny(36\%)}\\
\bottomrule
\end{tabular}
\caption{ORM training data for ProverQA dataset using Qwen.}
\label{tab:qwen-proverqa}
\end{table}

\begin{table}[ht]
\centering
\small
\setlength{\tabcolsep}{4pt} 
\begin{tabular}{llccc}
\toprule
\textbf{Mode} & \textbf{Echo} & \textbf{Total} & \textbf{Correct} & \textbf{Incorrect} \\
\midrule
CoT 0-shot & -- & 10009 & 7383 \raisebox{0.35ex}{\tiny(74\%)} & 2626 \raisebox{0.35ex}{\tiny(26\%)} \\
\midrule
\multirow{3}{*}{Echo 0-shot} 
  & True      & 10010 & 6552 \raisebox{0.35ex}{\tiny(65\%)} & 3458 \raisebox{0.35ex}{\tiny(35\%)} \\
  & False     & 10010 & 5459 \raisebox{0.35ex}{\tiny(55\%)} & 4551 \raisebox{0.35ex}{\tiny(45\%)} \\
  & Uncertain & 10004 & 4918 \raisebox{0.35ex}{\tiny(49\%)} & 5086 \raisebox{0.35ex}{\tiny(51\%)} \\
\midrule
    Echo-CoT & -- & 19105 & 7383 \raisebox{0.35ex}{\tiny(39\%)}& 11722 \raisebox{0.35ex}{\tiny(61\%)}\\
\bottomrule
\end{tabular}
\caption{ORM training data for FOLIO dataset using GPT4o.}
\label{tab:folio_datagen}
\end{table}

\begin{table}[htbp]
\centering
\small
\setlength{\tabcolsep}{4pt} 
\begin{tabular}{llccc}
\toprule
\textbf{Mode} & \textbf{Echo} & \textbf{Total} & \textbf{Correct} & \textbf{Incorrect} \\
\midrule
CoT 0-shot & -- & 39998 & 34486 \raisebox{0.35ex}{\tiny(86\%)} & 5512 \raisebox{0.35ex}{\tiny(14\%)} \\
\midrule
\multirow{3}{*}{Echo 0-shot} 
  & True     &40000 & 32865 \raisebox{0.35ex}{\tiny(82\%)} & 7135 \raisebox{0.35ex}{\tiny(18\%)}  \\
  & False&39996& 25725 \raisebox{0.35ex}{\tiny(64\%)} & 14271 \raisebox{0.35ex}{\tiny(36\%)}  \\
  & Uncertain &39968& 19385 \raisebox{0.35ex}{\tiny(49\%)} & 20583 \raisebox{0.35ex}{\tiny(51\%)}  \\
  \midrule
    Echo-CoT & -- & 63278 & 34486 \raisebox{0.35ex}{\tiny(54\%)} & 28792 \raisebox{0.35ex}{\tiny(46\%)} \\
\bottomrule
\end{tabular}
\caption{ORM training data for ProverQA dataset using GPT4o.}
\label{tab:proverqa_datagen}
\end{table}

\begin{table}[htbp]
\centering
\small
\setlength{\tabcolsep}{4pt} 
\begin{tabular}{llccc}
\toprule
\textbf{Mode} & \textbf{Echo} & \textbf{Total} & \textbf{Correct} & \textbf{Incorrect} \\
\midrule
CoT 0-shot & -- & 39197 & 27918 \raisebox{0.35ex}{\tiny(74\%)} & 11279 \raisebox{0.35ex}{\tiny(26\%)} \\
\midrule
\multirow{3}{*}{Echo 0-shot} 
  & True     &39199 & 21575 \raisebox{0.35ex}{\tiny(55\%)} & 17624 \raisebox{0.35ex}{\tiny(45\%)}  \\
  & False&39199& 20182 \raisebox{0.35ex}{\tiny(51\%)} & 19017 \raisebox{0.35ex}{\tiny(49\%)}  \\
  & Uncertain &39188& 14407 \raisebox{0.35ex}{\tiny(37\%)} & 24781 \raisebox{0.35ex}{\tiny(63\%)}  \\
  \midrule
    Echo-CoT & -- & 49197 & 27918 \raisebox{0.35ex}{\tiny(40\%)} & 41279 \raisebox{0.35ex}{\tiny(60\%)} \\
\bottomrule
\end{tabular}
\caption{ORM training data for JustLogic dataset using GPT4o. 10,000 records are sampled from the total echo records to preserve the final distribution of correct and incorrect values. }
\label{tab:justlogic_datagen}
\end{table}

\section{Resampling of Echo Data}
\label{app:sampling}
Unlike ProverQA and FOLIO, GPT-4o produces a significantly larger number of echoed rationales in the JustLogic dataset (Table~\ref{tab:justlogic_datagen}), resulting in a pronounced imbalance between correct and incorrect rationales. To address this, we adopt a weighted resampling approach that promotes both diversity and representational balance. Resampling is guided by two criteria: (1) BLEU-based diversity and (2) frequency of rationales per question.

To quantify the diversity of Echo-generated outputs relative to Chain-of-Thought (CoT) generations, we introduce a BLEU-based relative ranking metric. For each Echo response, we compute the BLEU score against its corresponding CoT responses. We then derive a percentile rank to assess the diversity of each Echo response within the group of candidates generated for the same question.

Formally, let the dataset comprise:

\begin{itemize}
    \item \textbf{Echo}: A set of generated hypotheses $\mathcal{H} = \{h_1, h_2, \ldots, h_N\}$
    \item \textbf{CoT}: A set of reference generations $R_i = \{r_{i1}, r_{i2}, \ldots, r_{iM}\}$ corresponding to the same input $x_i$ with sample size $M$.
\end{itemize}

For each Echo record $h_i \in \mathcal{H}$, where $h_i$ is associated with input $x_i$, compute the BLEU score $B_i$ using the set of CoT generations $R_i$ as references:
\[B_i = \text{BLEU}(h_i, R_i)\]

Let $\mathcal{G}_i \subseteq \mathcal{H}$ denote the set of all Echo hypotheses associated with a shared record identifier $i$ (i.e., same input $x_i$). Define the group-wise percentile rank of each BLEU score $B_i$ within group $\mathcal{G}_i$ as:
\[P_{\text{BLEU}, i} = 1 - \frac{\text{rank}(B_i \mid \mathcal{G}_i)}{|\mathcal{G}_i|}\]

where $\text{rank}(B_i \mid \mathcal{G}_i)$ is the ascending rank of $B_i$ within group $\mathcal{G}_i$, and $|\mathcal{G}i|$ is the group size. A higher $P{\text{BLEU}, i} \in [0, 1]$ indicates greater diversity relative to other hypotheses in the same group.

To reduce the sampling bias introduced by over-represented questions, we apply a frequency-based weighting scheme. For each input $i$, the frequency weight is defined as:
\[
W_{\text{freq}, i} = 1 - \frac{f_i - f_{\min}}{f_{\max} - f_{\min}}
\]
where $f_i$ is the frequency of a given record $i$, and $f_{\min}, f_{\max}$ denote the minimum and maximum frequencies across all records, respectively.This penalizes over-represented records during sampling.

The final sampling weight $w_i$ is a linear combination of diversity and frequency-based weights:
\[
w_i = \alpha \cdot P_{\text{BLEU}, i} + \beta \cdot W_{\text{freq}, i}
\]
where $\alpha$ and $\beta$ are hyperparameters controlling the relative importance of each component. We set $\alpha = 0.8$ and $\beta = 0.2$, and sample 10,000 hypotheses from the full set of 40,000.

\begin{figure*}[ht]
  \centering
  \includegraphics[width=\linewidth, trim=10 205 10 170, clip]{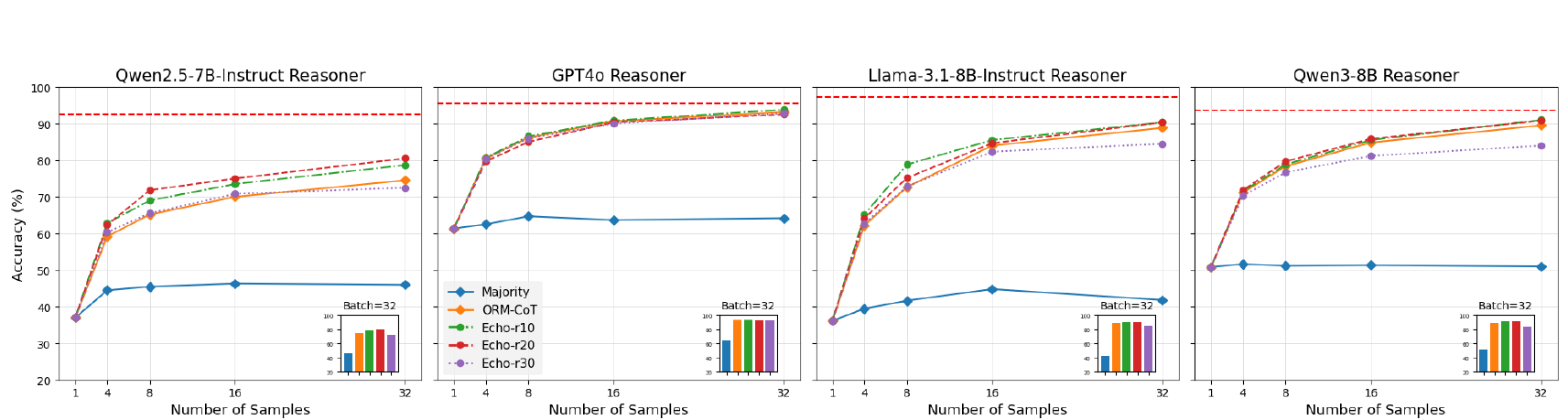}
  \caption{Performance of the ORM trained on GPT-4o-generated Echoed data with varying sample sizes. Sampling 10k examples yields the best performance, reflecting a balance between diversity and quality. Increasing the sample size to 30k degrades performance, likely due to the inclusion of lower-quality or redundant samples.}
  \label{fig:samplesize}
\end{figure*}

\paragraph{Sample Size Analysis.}
To justify the selection of 10,000 samples, we conduct an empirical analysis using subsets of 10k, 20k, and 30k examples, each drawn via the proposed weighted sampling procedure. As shown in Figure~\ref{fig:samplesize}, performance degrades with larger sample sizes due to the increased inclusion of low-quality or redundant hypotheses. This validates our choice of 10k as a balanced point between coverage and quality. 

\section{Ablation Studies}
\label{sec:ablation}
\begin{figure}[t]
  \centering
  \includegraphics[width=\linewidth]{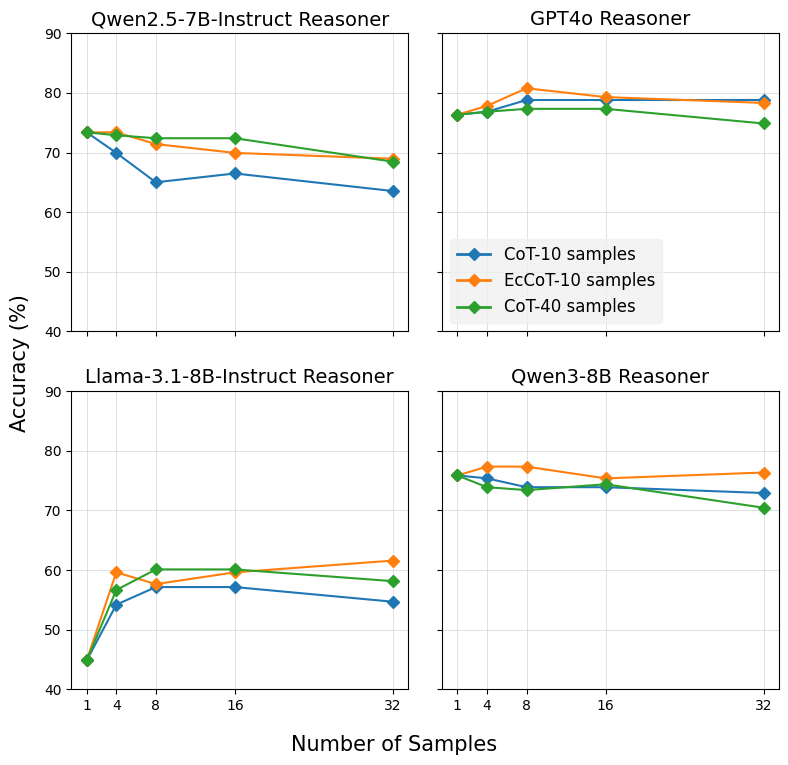}
  \caption{Performance of FOLIO with 10 samples using Echo and CoT vs 40 samples using simple CoT.}
  \label{fig:ormsamples}
\end{figure}

\paragraph{Sample Size of CoT vs EcCoT}
We aimed to measure the effect of sample size as determined by the number of training samples used. For FOLIO, we initially selected 10 samples per example, resulting in an Echo configuration of 10 samples per label. To determine whether the performance gains were due to the Echo method or simply the increased number of samples, we instead sampled 40 training examples using CoT for FOLIO and compared the performance of 10 Echo samples against 40 CoT samples. Figure~\ref{fig:ormsamples} shows that increasing the number of ORM samples does not outperform the Echo samples for both LLaMA and Qwen reasoners, where the performance was lower compared to CoT. This reinforces our motivation to use the Echo method rather than simply increasing the number of training records. We attribute this, in part, to the difference in diversity among the records, as discussed in detail in the Appendix~\ref{app:selfbleu}.

\begin{figure*}[ht]
  \centering
  \includegraphics[width=\linewidth, trim=110 70 110 70, clip]{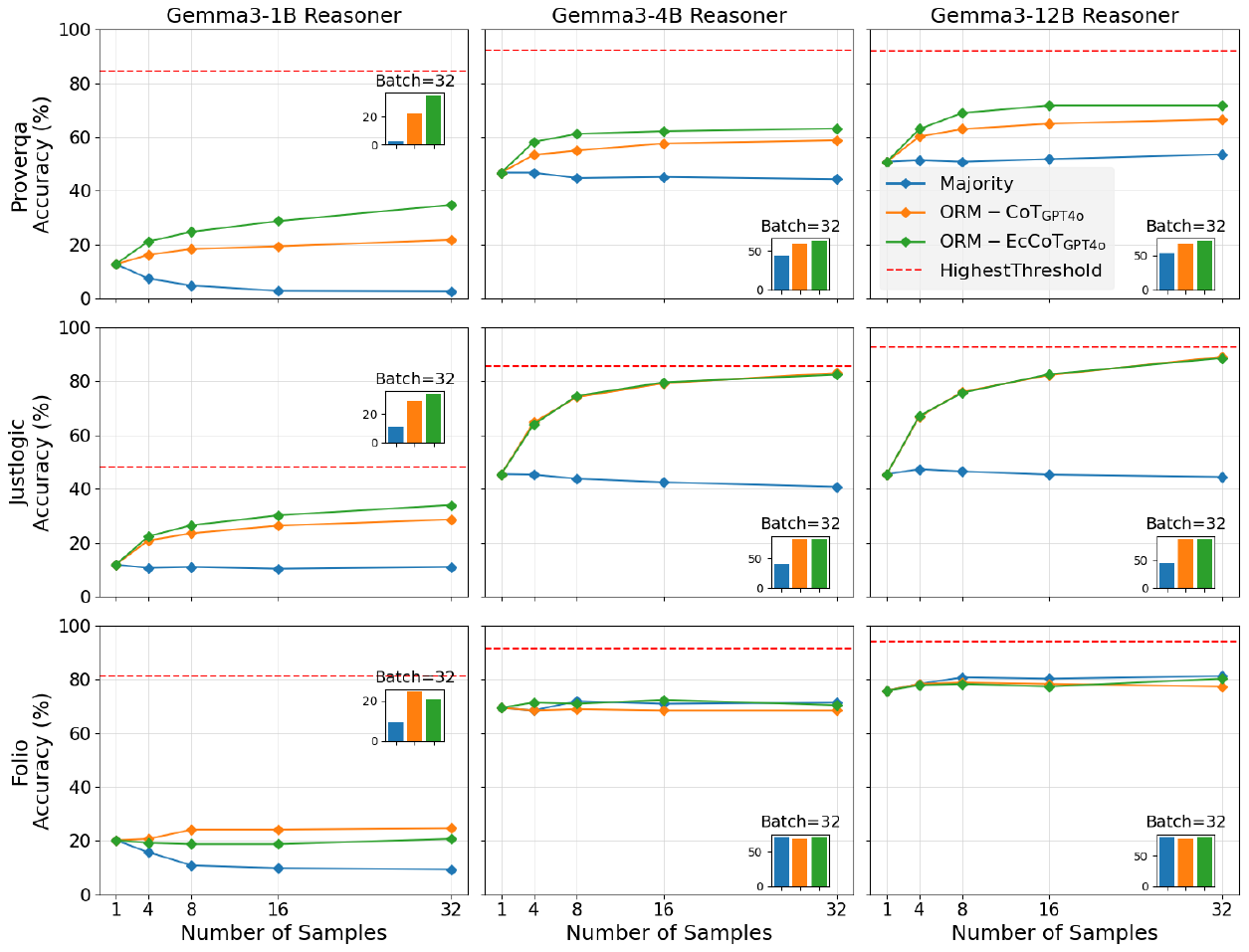}
  \caption{Performance of Gemma-based models of varying sizes on the ProverQA dataset shows a significant improvement with ORM, particularly benefiting the smaller models.}
  \label{fig:modelsize}
\end{figure*}

\paragraph{Impact of Reasoner Size}
We conducted an additional ablation study to examine the effect of LLM size on ORM performance. To this end, we evaluated Gemma models \cite{team2025gemma} of varying sizes (see Figure~\ref{fig:modelsize}). Notably, the smallest model, Gemma 1B, exhibited the most significant improvement—achieving nearly a 30\% increase in accuracy with the EcCot model compared to the 4B and 12B variants for ProverQA. These results highlight the effectiveness of using a simple CoT-based verification approach for smaller models, serving as a promising first step in validating reasoning paths.

\begin{table*}[ht]
\centering
\begin{tabular}{lcccc}
\hline \\
\textbf{Dataset} & \textbf{Sample Size} & \textbf{CoT Self-BLEU} & \textbf{EcCoT Self-BLEU} \\
\hline
ProverQA & 8 & 0.92 & 0.77 \\
FOLIO & 10 & 0.92 & 0.83 \\
FOLIO & 40 & 0.92 & 0.93 \\
\hline \\
\end{tabular}
\caption{Self-BLEU scores (lower is better) for standard CoT and EcCoT generations across datasets.}
\label{tab:selfbleu}
\end{table*}

\section{Echo Generation Diversity}
\label{app:selfbleu}
We evaluate generation diversity using self-BLEU scores, where lower values indicate higher diversity. Table~\ref{tab:selfbleu} reports average self-BLEU scores for both standard CoT and EcCoT generations. Across datasets, EcCoT consistently produces more diverse reasoning paths. Notably, increasing the number of standard CoT samples (e.g., from 10 to 40 in FOLIO) does not yield improvements in diversity.

\section{System Requirements for Experimentation}
\label{sys-req}
The Qwen and LLaMA models were accessed via the Hugging Face interface at \url{https://huggingface.co/Qwen/} and \url{https://huggingface.co/meta-llama/}, respectively. All models are gated and require access approval. GPT models were accessed through their API using batch calls to the /v1/chat/completions endpoint. Data generation, training, and inference were performed on a single A100 GPU, except for the API-based interactions.

\begin{table*}[ht]
\centering
\resizebox{\textwidth}{!}{%
\renewcommand{\arraystretch}{1.4}
\setlength{\tabcolsep}{8pt}
\begin{tabular}{|p{0.48\linewidth}|p{0.48\linewidth}|}
\hline
\textbf{Echoing Prompt:}{ Given the answer is Uncertain (C), please reason step by step, and put your final answer within \textbackslash\textbackslash  boxed\{\}.} & \textbf{CoT Prompt:}{Please reason step by step, and put your final answer within \textbackslash\textbackslash  boxed\{\}.} \\
\hline
\textbf{Input Context and Question:} \newline
William Dickinson was a British politician who sat in the House of Commons... William Dickinson supported the Portland Whigs. People who supported the Portland Whigs did not get a seat in the Parliament. \newline
\textbf{Question:} Is the statement "William Dickinson did not get a seat in Parliament" true, false, or uncertain?

\textbf{Answer Options:} A) True \quad B) False \quad C) Uncertain
&
\textbf{Input Context and Question:} \newline
William Dickinson was a British politician who sat in the House of Commons... William Dickinson supported the Portland Whigs. People who supported the Portland Whigs did not get a seat in the Parliament.
\newline
\textbf{Question:} Is the statement "William Dickinson did not get a seat in Parliament" true, false, or uncertain?
\newline
\textbf{Answer Options:} A) True \quad B) False \quad C) Uncertain
\\
\hline

\textbf{Reasoning:} \newline
1. William Dickinson sat in the House of Commons $\Rightarrow$ had a seat in Parliament. \newline
2. He supported the Portland Whigs $\Rightarrow$ such supporters didn’t get a seat. \newline
These are contradictory. Cannot resolve definitively. \newline
\textbf{Final Answer:} \textbackslash\textbackslash  boxed\{C\} (Uncertain)
&
\textbf{Reasoning:} \newline
1. William Dickinson supported the Portland Whigs. \newline
2. Portland Whigs' supporters did not get a seat. \newline
$\Rightarrow$ William Dickinson didn’t get a seat. \newline
\textbf{Final Answer:} \textbackslash\textbackslash  boxed\{A\} (True)
\\
\hline
\textbf{Predicted Answer:} C (Uncertain) \newline \textbf{Judgment:} Correct
&
\textbf{Predicted Answer:} A (True) \newline \textbf{Judgment:} Correct
\\
\hline
\end{tabular}}
\caption{Comparison of reasoning outcomes for an echoed prompt vs. a regular CoT.}
\label{tab:echo-vs-cot}
\end{table*}

\end{document}